\documentclass[conference]{preamble/IEEEtran}
\IEEEoverridecommandlockouts
\usepackage{cite}
\usepackage{amsmath,amssymb,amsfonts}
\usepackage{hyperref}
\usepackage{algorithmic}
\usepackage{graphicx}
\usepackage{textcomp}
\usepackage{xcolor}
\usepackage{caption}
\usepackage{listings}
\usepackage{multirow}
\usepackage{float}
\usepackage{subcaption}
\def\BibTeX{{\rm B\kern-.05em{\sc i\kern-.025em b}\kern-.08em
    T\kern-.1667em\lower.7ex\hbox{E}\kern-.125emX}}

\lstdefinelanguage[RISC-V]{Assembler}
{
  alsoletter={.}, 
  alsodigit={0x}, 
  morekeywords=[1]{ 
    lb, lh, lw, lbu, lhu, la,
    sb, sh, sw,
    sll, slli, srl, srli, sra, srai,
    add, addi, sub, lui, auipc,
    xor, xori, or, ori, and, andi,
    slt, slti, sltu, sltiu,
    beq, bne, blt, bge, bltu, bgeu,
    j, jr, jal, jalr, ret,
    scall, break, nop,
    mret, csrr, csrw
  },
  morekeywords=[2]{ 
    .align, .ascii, .asciiz, .byte, .data, .double, .extern,
    .float, .globl, .half, .kdata, .ktext, .set, .space, .text, .word
  },
  morekeywords=[3]{ 
    zero, ra, sp, gp, tp, s0, fp,
    t0, t1, t2, t3, t4, t5, t6,
    s1, s2, s3, s4, s5, s6, s7, s8, s9, s10, s11,
    a0, a1, a2, a3, a4, a5, a6, a7,
    ft0, ft1, ft2, ft3, ft4, ft5, ft6, ft7,
    fs0, fs1, fs2, fs3, fs4, fs5, fs6, fs7, fs8, fs9, fs10, fs11,
    fa0, fa1, fa2, fa3, fa4, fa5, fa6, fa7
  },
  morecomment=[l]{;},   
  morecomment=[l]{\#},  
  morestring=[b]",      
  morestring=[b]'       
}

\definecolor{mauve}{rgb}{0.58,0,0.82}

\begin{document}

\title{InTreeger: An End-to-End Framework for Integer-Only Decision Tree Inference\\
}
\author{\IEEEauthorblockN{Duncan Bart, Bruno Endres Forlin, Ana-Lucia Varbanescu, Marco Ottavi, Kuan-Hsun Chen}
\IEEEauthorblockA{
University of Twente, The Netherlands\\
\{d.bart, b.endresforlin, a.l.varbanescu, m.ottavi, k.h.chen\}@utwente.nl
}
}

\maketitle

\begin{abstract}
Integer quantization has emerged as a critical technique to facilitate deployment on resource-constrained devices. Although they do reduce the complexity of the learning models, their inference performance is often prone to quantization-induced errors. 
To this end, we introduce \textit{InTreeger}: an end-to-end framework that takes a training dataset as input, and outputs an architecture-agnostic integer-only C implementation of tree-based machine learning model, without loss of precision.
This framework enables anyone, even those without prior experience in machine learning, to generate a highly optimized integer-only classification model that can run on any hardware simply by providing an input dataset and target variable.
We evaluated our generated implementations across three different architectures (ARM, x86, and RISC-V), resulting in significant improvements in inference latency. In addition, we show the energy efficiency compared to typical decision tree implementations that rely on floating-point arithmetic. The results underscore the advantages of integer-only inference, making it particularly suitable for energy- and area-constrained devices such as embedded systems and edge computing platforms, while also enabling the execution of decision trees on existing ultra-low power devices.
 
\end{abstract}


\pagestyle{plain}
\thispagestyle{plain}

\section{Introduction}

While considerable research has focused on optimizing the resource requirements of convolutional neural networks and transformers, many applications do not require the complexity and overhead of such deep learning models, particularly in resource-constrained environments. Tree based models are well-suited for a variety of tasks where deep learning is not necessarily superior~\cite{SHWARTZZIV202284}. 
This has led to increased interest in more lightweight and efficient alternatives, such as Decision Trees (DTs) and their ensembles~\cite{NEURIPS2018_14491b75, Pedretti2021, 9923840, 10.5555/3294996.3295074, 10.1145/3508019, 10153433, 10.1145/3704727}.

DTs have both a lower computational complexity and energy consumption compared to deep learning methods, while also having a simpler training process, needing less data to achieve good performance.
Random Forests (RFs) further enhance efficiency by combining multiple decision trees, leading to improved accuracy and robustness while maintaining low resource usage~\cite{ULLAH2023116614}. 
Their inherent simplicity and interpretability make them ideal for many scenarios, where low latency and energy efficiency are critical, like edge computing~\cite{inemo}.

In addition, DTs can be implemented with minimal architectural dependencies, making them versatile across various hardware platforms~\cite{BeniniPaper}. 
This flexibility is increasingly valuable as computing environments diversify. 
The emergence of architectures like RISC-V, big.LITTLE systems, and hardware/software co-designed platforms has blurred the lines between server- and edge-grade hardware. As a result, tools that can generate AI models efficiently for both application-grade devices and ultra-low power microcontrollers without Floating-Point Units (FPUs) are more appealing. These tools must be architecture-agnostic, ensuring broad compatibility and maximizing performance across diverse systems for future AI deployments.

In this work, we introduce \textit{InTreeger}, an end-to-end framework that trains a dataset, and returns an architecture-agnostic, integer-only if-else tree implementation of a tree-based model in native C. 
This approach eliminates the need for the engineer responsible for final product integration to add libraries, set compiler-specific flags, or meet any architectural requirements beyond standard freestanding C. 
Furthermore, by providing an integer-only solution, performance gains are possible in all systems while allowing any microcontroller to be employed on the inference task.
As a result, tree-based AI-powered applications can be deployed directly at the edge to benefit application domain experts without the complications of adapting libraries or optimizing custom code.
In a nutshell, the paper is organized as follows:
\begin{itemize}
    \item Section~\ref{sec:backgroundSOTA} reviews the state-of-the-art inference frameworks, as well as optimizations for decision trees.
    \item In Section~\ref{sec:integerDT}, we describe our end-to-end framework, InTreeger, for the optimized inference of decision trees in an architecture-agnostic fashion.
    \item In Section~\ref{sec:evaluation}, we compare the performance of our framework against the state of the art, and analyze the performance bottlenecks in three different architectures: ARMv7, RISC-V, and x86. We also provide energy measurements, that demonstrate how using InTreeger leads to decreased energy consumption.
\end{itemize}
The developed framework is ready to be publicly released. A repository is prepared here for inspection:
\url{https://anonymous.4open.science/r/tl2cgen-InTreeger/}

\section{Background}
\label{sec:backgroundSOTA}
\subsection{Decision Trees and Their Ensembles}
In this work, we consider DTs under a supervised learning setting, where the training is done on labeled input data. Each DT is a binary search tree, and each branch node of the DT compares a particular feature from the input vector against a trained threshold. 
The inference process begins at the root. Depending on the comparison result, the input is funneled down either the left or right branch. This cycle of comparison and branching repeats until reaching a leaf node, which provides the predicted outcome. In the case of classification, this is often expressed as probabilities for the different possible classes.

Several well-known ensemble models assembled by DTs have been proposed to increase the accuracy and mitigate the issue of overfitting via the diversity of weak learners (i.e., each DT), such as Random Forests (RFs)~\cite{10.1023/A:1010933404324}, Extremely Randomized Trees~\cite{Geurts2006} and Gradient Boosted Trees (GBTs)~\cite{10.1214/aos/1013203451}.
Throughout this paper, we consider the most popular ensembles, namely RFs, to present our framework. Nevertheless, our framework supports all existing tree-based classification models. RFs construct multiple DTs during training, each built on different subsets of the data and features. The inference procedure of RFs works by evaluating each DT separately, and averaging the output predictions, which is in the form of probabilities in most modern, well-known, training frameworks, such as \textit{scikit-learn}~\cite{scikit-learn} for Python, or the \textit{randomForest}~\cite{randomForestR} package for R.
Since this work focuses on the inference deployment, we omit the background of how DTs and RFs is trained in details. 



\subsection{Inference Optimizations and Deployment Frameworks}
Performance optimization of DTs and their ensembles is a widely studied topic~\cite{10.1145/3652032.3659366}. A popular example is the C\texttt{++} implementation for RFs by Wright and Ziegler~\cite{JSSv077i01}, which is widely used by statisticians. For machine-close programming languages like C and C\texttt{++}, the prominent concept of native trees, where nodes are processed throughout a narrow loop, and if-else trees, where nodes form nested if-else constructs, is introduced by Asadi et al.~\cite{6513227}. These two kinds of implementations result in very different execution behaviors regarding, for example, memory locality~\cite{DBLP:conf/icdm/BuschjagerCCM18}. For RFs, several implementations have been deployed on various hardware architectures, such as GPUs, FPGAs~\cite{6239820, 7962153} and content addressable memories~\cite{Pedretti2021, 10247695}.
Tabanelli et al. benchmark various RF implementations on a single-core RISC-V MCU, where an optimized native tree is proposed~\cite{BeniniPaper}. Similarly, Daghero et al. propose runtime optimizations for IoT devices which have dynamic energy budgets~\cite{10153433}. In this work, we focus on executing the DTs onto three different CPU architectures with application-level processors: ARM, x86 and RISC-V. 

To facilitate the deployment of inference models for application domain experts, several deployment frameworks have been developed.
Buschjäger et al. have maintained a code generator to derive an optimized implementation of RFs in C~\cite{DBLP:conf/icdm/BuschjagerCCM18}, hosting several accuracy-preserved optimizations at the programming level such as~\cite{10.1007/978-3-031-26419-1_32, DBLP:journals/tecs/ChenSHBLLMC22, 10546851}. Although the approaches mentioned above provide various optimized implementations of RFs based on different insights, quantization is a very popular step to deploy the inference models, especially for resource-constrained devices. 
In this work, we leverage the code generation module named \textit{tl2cgen}\footnote{\url{https://github.com/dmlc/tl2cgen}}, to realize our optimizations on top of the C implementation of RFs and GBTs based on the if-else tree, this module is a decoupled submodule from the well-known model compiler Treelite~\cite{DBLP:conf/mlsys/2018}. 




\subsection{Quantization}

The data types utilized in DTs are determined by the data type used during training. For instance, if the training dataset contains floating-point values, the resulting split values are often floating-point numbers as well. However, relying on floating-point numbers introduces performance overheads due to two primary reasons. Not only do floating-point operations generally take longer to process, managing floating-point data also incurs extra overhead due to transferring data to floating-point registers, and increased memory loads due to the use of more complex encoding schemes. Moreover, the requirement of FPUs diminishes the number of platforms that can support the inference, such as ultra-low power devices at the edge.

For deep learning, quantization has been extensively studied~\cite{DBLP:conf/cvpr/JacobKCZTHAK18}, while it is much less explored in the context of DTs. 
During the training of DTs, fixed-point quantization has been considered, e.g., by the means of histogram binning~\cite{XGBoost} or through quantized boosting~\cite{10.1007/978-3-030-46150-8_35}. However, the comparison thresholds generated by training frameworks, such as ~\cite{scikit-learn}, might still be floating-point numbers. Daghero et al. adopt a symmetric min-max quantizer to quantize the leaf probabilities, with the range of the values that the accumulated probabilities can assume to derive the quantizer parameters~\cite{10153433}. While the focused implementation employs native trees, this work targets if-else trees, which are more suitable for microcontrollers with limited RAM.

\subsection{FlInt}
\label{subsec:FlInt}

Hakert et al. propose reinterpreting floating-point numbers as bit vectors and leveraging integer arithmetic for comparisons~\cite{10546851}. Unlike typical quantization, this method maintains full accuracy and allows training to be done using floating-point arithmetic. 
The authors demonstrate this implementation on x86 and ARMv8 architectures. However, when relying on the compiler, different optimization strategies may be applied, each having varying impacts depending on the target platform. Moreover, the implementation is highly architecture-dependent, leading to different levels of success in achieving the desired performance gains across platforms.

\section{InTreeger Framework}\label{sec:integerDT}

Our framework, InTreeger, aims to bridge the gap between the existing machine learning libraries and ultra-low power devices, enabling the inference of tree-based models on these devices with minimal effort from the user, and without loss of accuracy. We achieve this by converting all floating-point operations in the inference of DTs to integer alternatives.

When converting floating-point values to integers, the most common approach is quantization. However, a key drawback of this method is the loss of accuracy it introduces. As shown above, FlInt addresses this issue by enabling threshold evaluations in the decision tree nodes using integer arithmetic. This solution can lead to speed improvements in certain scenarios, while also reducing the instruction count, which in turn minimizes the size of the resulting binary.

Despite these advantages, an essential limitation still remains. Namely, in the leaf nodes of the tree, the probabilities for each class are still represented as floating-point values. A code snippet is provided in \autoref{lst:float_probs}. Line 1 shows the threshold delivered by FlInt, but in lines 2-3, the class probabilities remain single precision floating-point numbers. This poses a problem for hardware that lacks floating-point support, such as ultra-low power devices, and limit the applicability of DTs. 

\begin{figure}[h]
 \vspace{-0.3cm}
\centering
\captionsetup{width=\columnwidth}  
\begin{lstlisting}[caption={
A C implementation of a leaf node using FlInt for the threshold and floats for the probabilities.}, label={lst:float_probs}] 
if ( !(data[9].missing != -1) || ((*( ((int*)(data)) + 9 ))<=((int)(0x3eb4dc48))) ) )
    result[0] += (float)0.2717557251908397;
    result[1] += (float)0.7282442748091603;
\end{lstlisting}
 \vspace{-0.6cm}
\end{figure}

\subsection{Probability to Integer Conversion} \label{sec:prob_to_int}

Generally, in an if-else tree implementation of an RF model, the probabilities from each DT in the ensemble are summed up and divided by the total number of trees in the ensemble, giving us the average of all probabilities for each class. These probabilities are all represented as 32-bit floating-point numbers. Given that probabilities always fall between 0 and 1, we represent them as integers by multiplying each probability by the maximum value for a 32 bit unsigned integer, $2^{32}$, and rounding down, saving the result as a 32 bit unsigned integer value. This transformation offers precision up to the 10th decimal place (more precisely, to $\frac{1}{2^{32}}$). By applying this transformation to all probabilities during the code generation phase, it eliminates the need for floating-point arithmetic during the inference. In essence, this is simply a fixed-point representation of the probability, with a scaling factor of $2^{32}$.

However, a direct summation of these integer-converted probabilities across the trees in the ensemble would result in integer overflow. 
To prevent this, we divide each converted probability by the number of trees in the ensemble at the time of code generation. These scaled values are then summed to obtain the final predicted probabilities. Since this division is performed during code generation, it has no effect on the inference process.
Thus, the final scaling factor for probabilities in the context of RFs is $\frac{2^{32}}{n}$, where $n$ is the number of trees in the ensemble. This changes the accuracy to $\frac{n}{2^{32}}$.

As an example, let us have an RF model, with two classes, and 10 trees. One of the leaf nodes assigns a 75\% chance to the first class, and a 25\% chance to the second. The floating-point version of a leaf node in an if-else tree implemented in C is as follows:

\begin{figure}[H]
 \vspace{-0.3cm}
\centering
\captionsetup{width=\columnwidth}  
\begin{lstlisting}[] 
result[0] += (float)0.75;
result[1] += (float)0.25;
\end{lstlisting}
 \vspace{-0.3cm}
\end{figure}

When converting to integers, we multiply these values by $\frac{2^{32}}{10}$, as there are 10 trees in the ensemble. This gives us the values $0.75\times\frac{2^{32}}{10}= 322122547.2$, which is rounded down to 322122547, and $0.25\times\frac{2^{32}}{10}= 107374182.4$, which is rounded down to 107374182. This results in the following code:

\begin{figure}[H]
 \vspace{-0.3cm}
\centering
\captionsetup{width=\columnwidth}  
\begin{lstlisting}[] 
result[0] += 322122547;
result[1] += 107374182;
\end{lstlisting}
 \vspace{-0.3cm}
\end{figure}

This transformation will not impact the accuracy performance of the RF model in any realistic scenario. There are, however, two edge-case scenario's where this conversion will lead to a loss of accuracy.

An IEEE 754 single-precision float has a precision of 24 bits \cite{8766229}, which results in an accuracy of $\frac{1}{2^{24}}$, or about 7 significant digits. Thus, when the probability is in the range $[0.1, 1)$, we have a precision of 7 decimal places. However, since a floating-point number also has an exponent, if the probability is in the range $[0.01, 0.1)$, the precision improves to 8 decimal places, for the range $[0.001, 0.01)$, the precision is 9 decimal places, and so forth.

In the best case scenario, our fixed-point representation is accurate up to 10 decimal points. Thus, we lose accuracy if the floating-point representation is more accurate than the fixed-point representation. This happens when the probability is smaller than approximately 0.001. However, in this case, the fixed-point representation is still accurate until the 10th decimal. Even though a floating-point representation would in theory be more accurate, most real-world scenario's do not require such precision.

The second edge-case arises when the number of trees in the ensemble is so high, that the decreased fixed-point scaling factor causes the accuracy to be lower than the floating-point accuracy. This happens when 

$$\frac{n}{2^{32}} > \frac{1}{2^{24}},$$

\noindent
which simplifies to $n>256$. Thus, if there are more than 256 trees in the forest, single precision floating-point numbers are more accurate. However, as the authors of \cite{oshiro2012many} demonstrate, there is no significant accuracy improvement from 128 trees onwards in the ensemble. 


\subsection{The End-to-End Framework}
We choose to implement both FlInt and our probability to integer conversion in the existing model compiler \textit{tl2cgen} for several reasons. \textit{Tl2cgen} is a well-established framework, with an existing community. This allows these optimizations to be easily adopted by the community. Furthermore, it already implements a variety of other optimizations, improving the efficiency of the generated C code. Finally, \textit{tl2cgen} offers support for various training frameworks, such as \textit{XGBoost}, \textit{LightGBM}, and \textit{scikit-learn}, which makes our integer-only implementation easily available to a wide range of existing applications with existing models.

With the integer-only model compiler in place, we created a user-friendly, easy to use framework, that trains an RF model on a given dataset, and compiles it into a highly optimized integer-only C program, which can be compiled to any architecture.
Next to generating C code, this framework also allows the user to directly compile the generated C code, and use it as a Python predictor function.
InTreeger leverages existing, well-established libraries. 

\begin{figure}[t]
    \centering
    \includegraphics[width=0.9\linewidth]{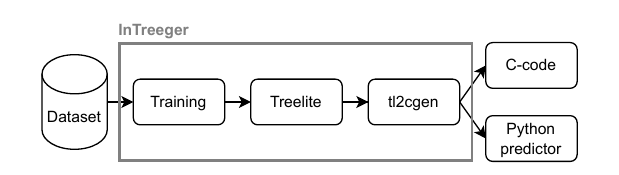}
    \caption{The overview of InTreeger. From input dataset to integer-only architecture-agnostic if-else tree.}
    \label{fig:pipeline}
\end{figure}

A schematic overview of the underlying pipeline of the framework is shown in \autoref{fig:pipeline}.
The pipeline begins with the preprocessed training dataset, which must be annotated with the corresponding classes, as is the case with any supervised classification algorithm. This annotated dataset is then passed into a Python RF training library of choice, such as \textit{XGBoost}, \textit{LightGBM}, and \textit{scikit-learn}, which will output a Python object of the trained model.

Next, this model object, regardless of the library used, is converted into a common \textit{Treelite} representation. This serves as a standardized intermediary that simplifies subsequent processing and optimization of the model.
With the updated \textit{tl2cgen} library, extended to support FlInt and our probability to integer transformation, we can generate a standalone C file. This file contains an integer-only if-else implementation of the trained RF model, which can be deployed on any architecture.
Alternatively, if a Python workflow is preferred, the optimized model can be compiled directly, allowing for the use of the predict function within Python directly, which is handy for application domain experts.

One of the main benefits of InTreeger is that it has the potential to save application domain experts a lot of time and effort, as it automates processes that can be quite time-consuming for people without experience dealing with various tools, such as finding the right training methods, fixing dependencies, or optimizing the inference process. The generated C code does not use any libraries except for a few standard libraries, so no other installation is required to run the generated models on any platform.


\begin{table}[]
\caption{The experiments were performed using the following cores and architectures.}
\label{tab:cores}
\resizebox{\columnwidth}{!}{
\begin{tabular}{l|l|r|r|l}
Core                                                         & ISA        & Word Size & Frequency               & Memory Hierarchy                                                                                                     \\ \hline
EPYC 7282                                                    & x86        & 64        & 2.8GHz                  & \begin{tabular}[c]{@{}l@{}}32KB I-Cache / 32KB D-Cache per core\\ 512 kB L2-Cache / 16 MB L3-Cache\end{tabular}      \\ \hline
Cortex-A72                                  & ARMv7      & 32        & 1.8GHz & \begin{tabular}[c]{@{}l@{}}48KB I-Cache / 32KB D-Cache per core\\ 1 MB Shared L2-Cache\end{tabular} \\ \hline
U74-MC                                                       & RV64IMAFDC & 64        & 1.2 GHz                 & \begin{tabular}[c]{@{}l@{}}32KB I-Cache / 32KB D-Cache per core\\ 2 MB Coherent Banked L2-Cache\end{tabular}         \\ \hline
FE310 & RV32IMAC      & 32        & 16 MHz                 & 32 MB External Flash  / 16 DTIM KiB                                                                                                 
\end{tabular}}
 \vspace{-0.3cm}
\end{table}

\section{Empirical Evaluation}
\label{sec:evaluation}
This section examines both the accuracy of InTreeger across various datasets and its effectiveness in terms of performance and portability across different architectures and cores.
We demonstrate that the effectiveness of the technique is directly related to how well the immediate value can be mapped to instructions in the binary, which is directly affected by the chosen architecture. For this, we select the cores shown in \autoref{tab:cores}. In details, we evaluate experimentally the following aspects:

\begin{itemize}
    \item The accuracy loss of InTreeger for different datasets.
    \item The effectiveness of mapping the immediate fields (instruction conversion) for each architecture.
    \item Performance and instruction count for floating-point, FlInt, and InTreeger implementations on various cores and architectures.
\end{itemize}

Runtime performance is measured using the perf utility, with 10,000 function call replications to enhance runtime contribution and benchmark representativeness. All experiments use the -O3 compiler flag for realistic real-world performance.

While the ARM core can execute ARMv7 instructions in compatibility mode, which might not yield optimal performance for that architecture, we show in this section that both the architecture and the way the code is compiled are critical factors influencing performance.
Additionally, we demonstrate the framework's flexibility through a use case, evaluating inference code on the SiFive FE310 SoC, a real-world RISC-V microcontroller.

Finally, we provide energy measurements, that show how using InTreeger will yield significant energy savings in a real-world unoptimized scenario. For these measurements, we used the Joulescope JS220 platform~\cite{joulescope_js220}. The Joulescope JS220 is a precision energy measurement tool designed for accurate profiling of current, voltage, and power in electronic devices. It supports dynamic range measurements, spanning from nanoamps to amps, making it ideal for low-power embedded systems. 

\subsection{Datasets}
For all experiments throughout this section, we make use of two distinct datasets. 
The first dataset is the \textit{Statlog (Shuttle)} dataset (referred to as the Shuttle dataset) from the UC Irvine Machine Learning Repository \cite{shuttle_dataset}. This dataset has 58000 instances, 7 features, and 7 classes.

The second dataset is the \textit{ESA Anomaly Dataset} (referred to as the ESA dataset) from the European Space Agency \cite{esa_dataset}. From this dataset, we use the first three months of data, which contains 262081 instances and 87 features. We define the target value to be 1 if there is an anomaly in any of the features, and 0 if this is not the case, resulting in 2 classes.
We opted for two datasets related to edge machine learning in space, as this is a typical scenario where minimizing computing requirements is a necessity due to power constraints.
The datasets do, however, vary greatly in size, number of features, and number of classes. Therefore, we deem the obtained results to be generalizable to other datasets.

\begin{figure}[t]
    \centering
    \includegraphics[width=0.9\linewidth]{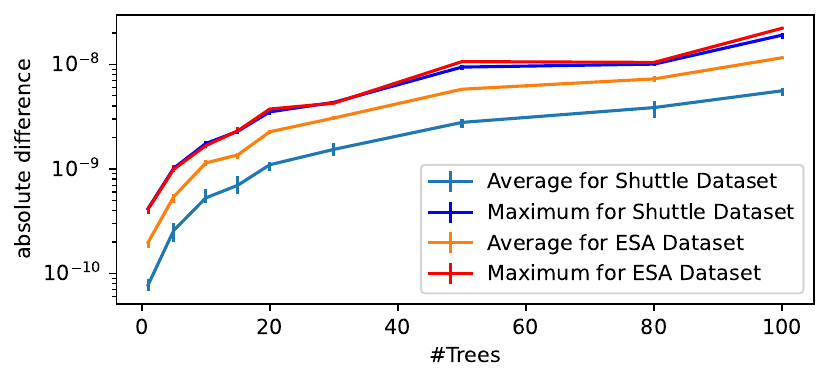}
    \caption{The differences between the observed probabilities for the standard floating-point implementation, and our integer-only implementation.}
    \label{fig:accuracy}
     \vspace{-0.5cm}
\end{figure}
\subsection{Accuracy}
To verify the statement that converting the probabilities to unsigned integers does not impact the accuracy of the models, we ran extensive tests on our datasets, observing the difference between the prediction accuracy of the original model and the integer-only model.

We measured this difference by making a 75\%-25\% split on the datasets, and using the 75\% part to train the model using scikit-learn. We then convert this trained model to two different if-else tree implementations: 1) one default implementation, using 32 bit floating-point probabilities, and 2) our method with the probability to integer conversion.

Both models were used to make predictions over the 25\% test set. As hypothesized, the predictions of both models were identical on every data sample we tested. To verify the validity of our results, we ran this test on 10 different randomized train-test splits, and for RF models with up to 100 trees, and observed identical results every time.

Furthermore, we observed the difference between the predicted probabilities for each class of the two datasets. The results of these experiments are shown in \autoref{fig:accuracy}. These results align with the concept explained in Section \ref{sec:prob_to_int}, where the maximum differences are proportional to the number of trees in the ensemble, and are in the order of magnitude of $10^{-10}$ for a single DT, and around $10^{-8}$ for an RF model with 100 trees. The average differences are slightly below that. As we experimentally demonstrate, these tiny differences have a negligible effect on the accuracy of the RF models.

\subsection{Immediate Conversion}


InTreeger's immediate conversion technique is an extension from FlInt~\cite{10546851}. The FlInt paper demonstrated how the conversion can happen to x86 and ARMv8 architectures, but it does not cover the architectures that would most likely be found on low-energy edge devices.
Therefore, in this section, we experimentally demonstrate the differences that result from the compiler's optimization in the InTreeger implementation in both RISC-V and ARMv7.

\autoref{lst:intreeger} and \autoref{lst:intreegerARM} demonstrate this variation for RISC-V and ARMv7, respectively. The RISC-V implementation (both 32 bits and 64 bits) can encode up to 20-bits into the immediate field of the \textit{lui} instruction, as shown in line 3 of \autoref{lst:intreeger}. This is a natural fit for the conversion applied by FlInt, as it only uses the upper bits to represent the Split Value (SV)~\cite{10546851}. For the probability encoding, the compiler will either use a single \textit{lui} instruction or a \textit{lui} followed by an \textit{add} instruction to calculate the lower 12 bits, this can be seen in lines 8 and 9 of \autoref{lst:intreeger}.

\begin{figure}[t]
\centering
\captionsetup{width=\columnwidth}  
\begin{lstlisting}[caption={InTreeger implementation in RISC-V.}, label={lst:intreeger}] 
#if ((*(((int*)(data)) + 5 ))<=((int)(0x42af0000))) 
lw      a4,20(a0)   # Load data
lui     a5,0x42af0  # Get immediate upper 20 bits
blt     a5,a4,23e0c # Branch

# result[0] += 4292021501;
lw      a3,0(a2)    # Load result[0]
lui     a0,0xffd31  # Get immediate upper 20 bits
addiw   a0,a0,-771  # Calculate lower 12 bits
addw    a3,a3,a0    # Add to result[0]
sw      a3,0(a2)    # Store result[0]

\end{lstlisting}
 \vspace{-0.5cm}
\end{figure}
\begin{figure}[t]
\centering
\captionsetup{width=\columnwidth}  
\begin{lstlisting}[caption={InTreeger implementation in ARMv7.}, label={lst:intreegerARM}] 
# First comparison:
# if ((*(((int*)(data)) + 5))<=((int)(0x42af0000)))
ldr	r3, [pc, #744] # Load 0x42af0000 from a pc offset 

# Second comparison:
# if ((*(((int*)(data)) + 2))<=((int)(0x428a0000)))
ldr	r1, [r0, #8]      # Load data
sub	r3, r3, #2424832  # Calculate new immediate
cmp	r1, r3            # Compare r1 and r3
bgt	1a650             # Branch

# result[0] += 4292021501;
ldr lr, [r2]         # Load result[0]
ldr r3, [pc, #320]   # Load 4292021501 from a pc offset
add r3, lr, r3       # Add r3 into result[0]
str r3, [r2]         # Store result[0]

\end{lstlisting}
 \vspace{-0.5cm}
\end{figure}

\begin{figure*}[h!]
    \centering
    \includegraphics[width=\textwidth]{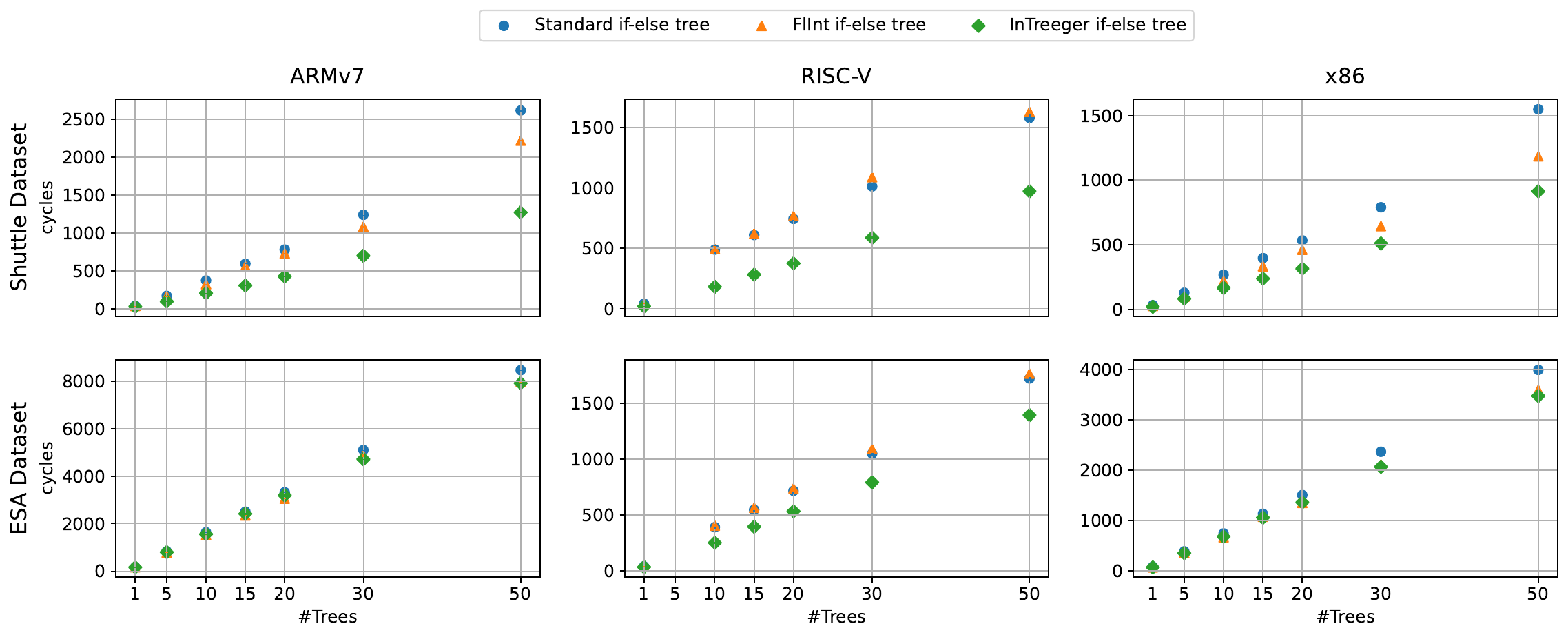}
    \caption{Performance evaluation (elapsed cycles) of the inference for the two datasets over the selected architectures and application-level systems with a variable number of trees in each ensemble. }
    \label{fig:arch-results}
\end{figure*}

ARMv7 (and ARMv8), on the other hand, does not provide an analogous \textit{lui} instruction. Loading of large immediates in ARM is usually done by referencing the program counter, as these values are embedded into instruction data. Line 3 of \autoref{lst:intreegerARM} shows that for the FlInt optimization, the compiler can choose to load the first SV from memory, and then use it as reference to calculate the following SVs (line 8). This behavior is also seen when calculating the probabilities, with the immediates being loaded from program memory as seen in \autoref{lst:intreegerARM} line 14. This makes ARMv7 have a slightly higher register and memory footprint than the RISC-V implementation.

We demonstrate the float-only implementation in \autoref{lst:naive} in RISC-V. From an instruction count point of view, we see that all implementations will have mostly the same number of instructions, with the RISC-V implementation possibly having 7 instructions depending on the probability value. However, the key is to notice that the required operations for the InTreeger implementation are arguably the most basic operations in a pipeline for both architectures (load, add, store, compare). In contrast, the floating-point operations in the naive and FlInt implementations may exhibit higher latency, depending on how they are implemented. This can be directly related to the cost of the FPU implementation in hardware. This factor has a direct energy and performance trade-off that is easily exploitable, as we will see in the following section.

\begin{figure}[t]
\centering
\captionsetup{width=\columnwidth}  
\begin{lstlisting}[caption={Naive (float) implementation in RISC-V.}, label={lst:naive}]  
#if (data[5].fvalue <= (float)87.5))
fmv.w.x  ft2,a5      # Move data[5].fvalue into ft2 
flw      fa2,488(gp) # Load 87.5 info fp register 
fle.s    a5,ft2,fa2  # a5 = (ft2 <= fa2)
bnez     a5,23de0    # Check a5

#result[1] += 0.013623978201634877;
flw      fa4,4(a2)   # Load result[1]
fld      fa5,272(gp) # Load probability
fadd.s   fa4,fa4,fa5 # Add probability
fsw      fa4,4(a2)   # Store result[1]

\end{lstlisting}
 \vspace{-0.5cm}
\end{figure}

\subsection{Architecture Performance}

The results for the architecture evaluation can be seen in \autoref{fig:arch-results}. The results show that the standard floating-point method has the worst performance in all cores, with FlInt's performance being quite close for the ARMv7 and RISC-V architectures. InTreeger performed the best in all cases we tested. The exact performance gains depend first of all on the number of trees in the ensemble, but more noticeably on the number of classes of the dataset. We see that the Shuttle dataset, which has 7 classes, shows much larger performance gains than the ESA dataset, which has only 2 classes. This can be explained by the fact that in each leaf node, we add a probability to each class of the dataset. Thus, the performance gains caused by the conversion scales linearly with the number of classes in the dataset.

In the best case, which is the Shuttle dataset on the ARMv7 architecture with 50 estimators, InTreeger's speedup was 2.1x, or a runtime reduction of 52\%. The worst performance was for the ESA dataset on the ARMv7 architecture, which still saw an average runtime reduction of 4.8\%.

The performance gains achieved by InTreeger can not simply be explained by only a slight reduction in instruction count. Most likely the compiler has an easier time using register optimizations, limiting memory accesses, while not having to handle two register files simultaneously. Since RISC-V and x86 have better dedicated instructions to immediate handling than ARMv7, the performance gains are more significant.


\subsection{Use Case: RISC-V Microcontroller}

To demonstrate that InTreeger can be applied across various architectures, scenarios, and computing capabilities, we implemented an RF model using the Shuttle dataset, with 30 trees and a maximum depth of 5, on a small RISC-V microcontroller. The chosen SoC is the SiFive FE310, integrated on the SparkFun RED-V RedBoard, operating at a clock frequency of 16 MHz. The generated code can be directly added as the system's entry point, or called as a function by the firmware. In our case, the firmware contained the necessary instrumentation to read out the performance counters for evaluation.

After compilation, we achieve a memory utilization of 42,382 bytes for the text section, 8 bytes for the data section, and 1,152 bytes for the BSS section, resulting in a total memory footprint of 43,542 bytes. The compilation used the \texttt{-march=rv32imac zicsr zifencei -mabi=ilp32} flags, ensuring compatibility with the FE310 SoC. This type of model fits well on devices with a large amount of read-only memory (QSPI flash) and limited RAM.

The inference was repeated 10,000 times in the firmware to obtain representative values, achieving an execution time of 0.6 seconds per inference, which corresponds to a rate of 1.66 inferences per second. During execution, the system performed 7,243,185 instructions per inference, with 0.746 instructions per cycle (IPC). This result is primarily due to the longer fetch time (i.e., worst case up to 24 cycles) for uncached instructions from the QSPI.

\subsection{Energy Consumption}

\begin{figure}[h]
    \centering
    \includegraphics[width=\linewidth]{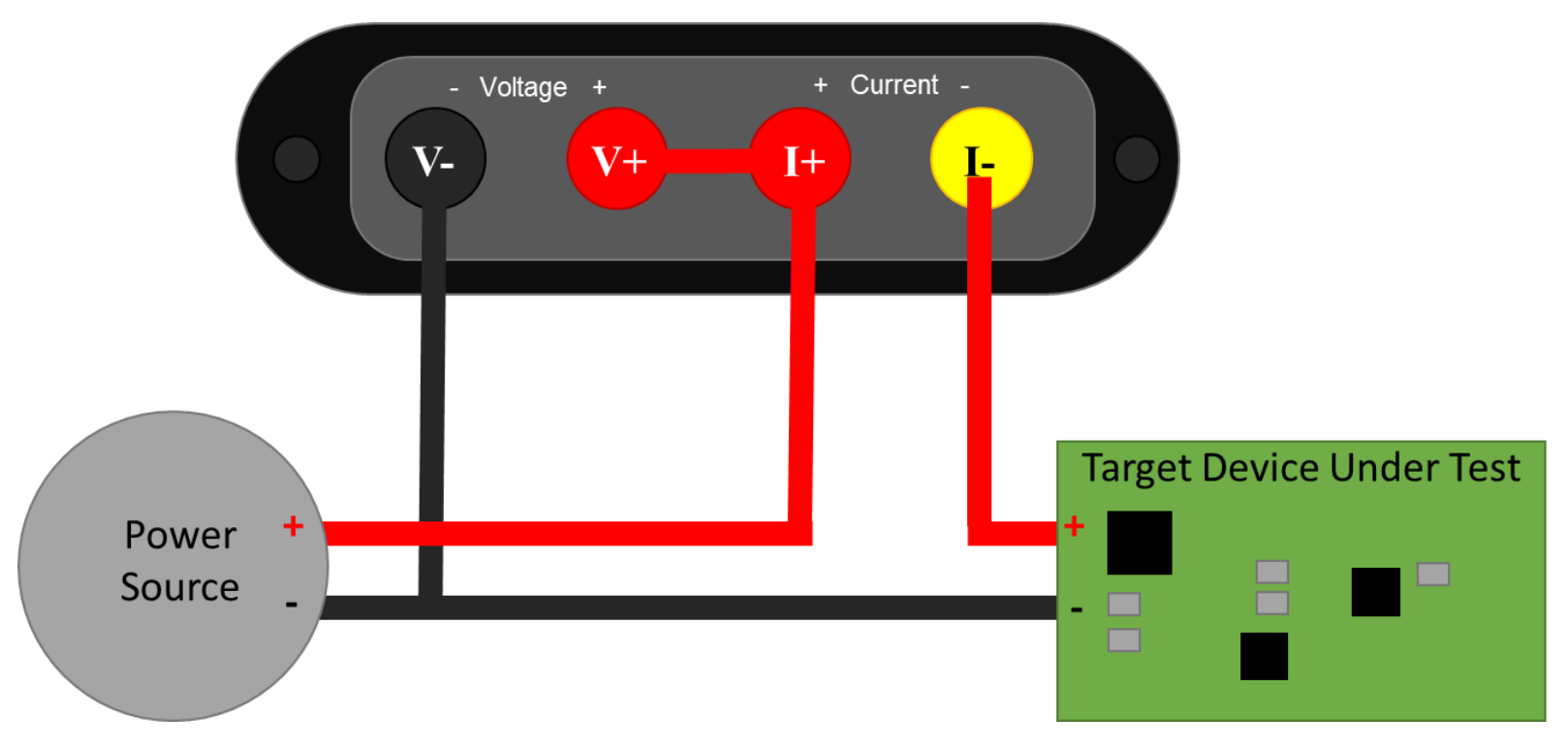}
    \caption{A schematic overview of the setup that was used for conducting the power measurements. Extracted from  \cite{joulescope_man}.}
    \label{fig:power_setup}
\end{figure}

A schematic overview of the setup used for the power measurements is shown in \autoref{fig:power_setup}. The target device under test for the experiments conducted in this section is the ARMv7 core which is also used for the performance evaluation.

On this setup, we run 14,500,000 inferences of a Random Forest model of the Shuttle dataset with 50 Decision Trees and a maximum depth of 7 on both the standard floating-point implementation, as well as our integer-only implementation. The number of inferences stemmed from the number of rows in the test set of the Shuttle dataset, 14,500, which we multiplied by 1000 to artificially create a longer runtime to increase the accuracy of our measurements.

\begin{figure}[h]
\begin{subfigure}{\linewidth}
\includegraphics[trim=70 30 90 60,clip,width=\textwidth]{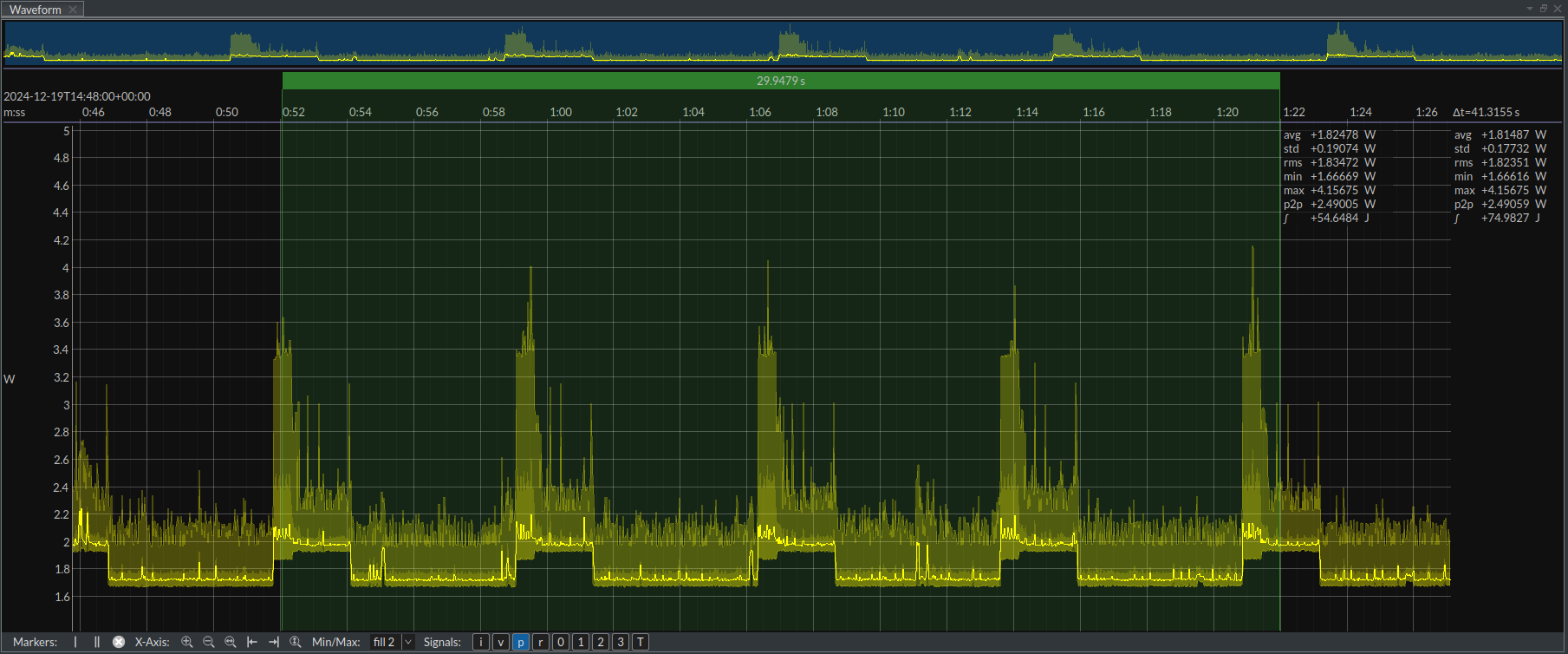}
\caption{Baseline}
\label{fig:power_baseline}
\end{subfigure}

\bigskip

\begin{subfigure}{\linewidth}
\includegraphics[trim=70 30 90 60,clip,width=\textwidth]{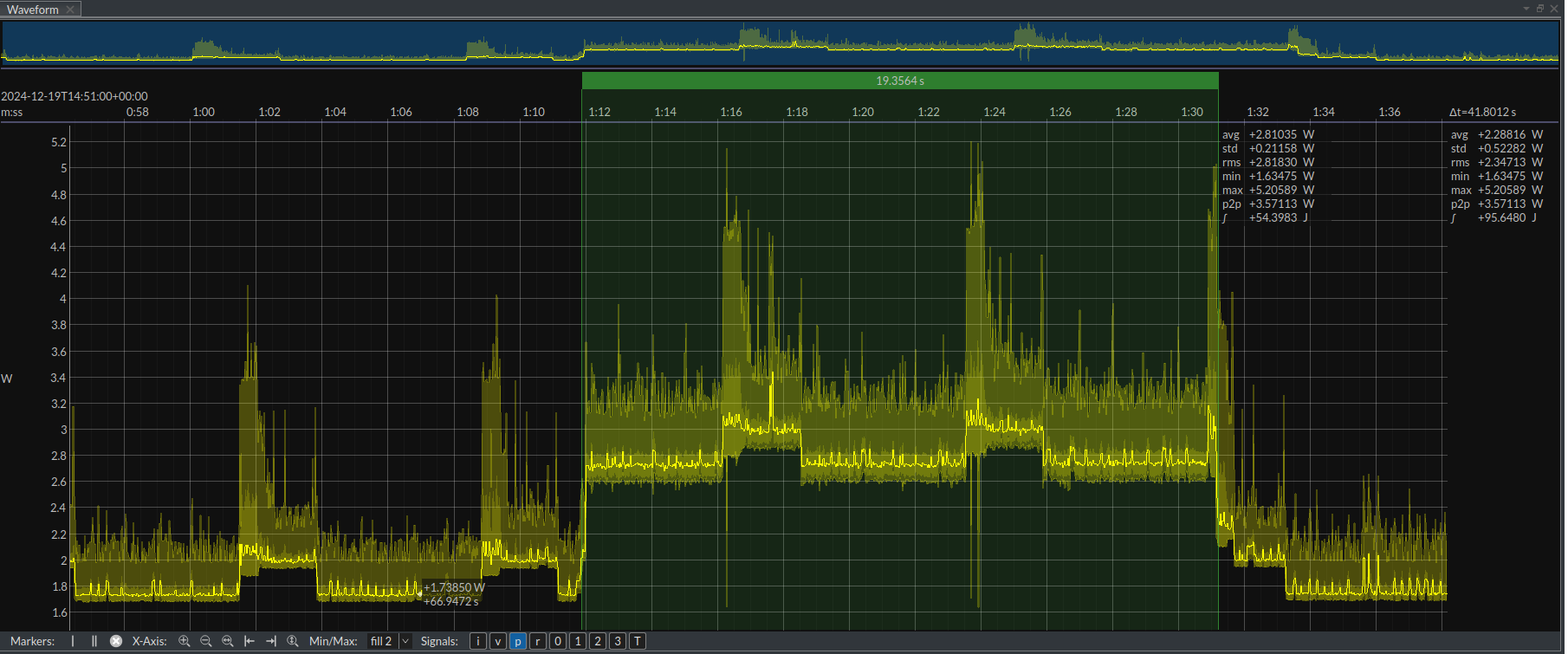}
\caption{Floating-point implementation}
\label{fig:power_float}
\end{subfigure}

\bigskip

\begin{subfigure}{\linewidth}
\includegraphics[trim=70 30 90 60,clip,width=\textwidth]{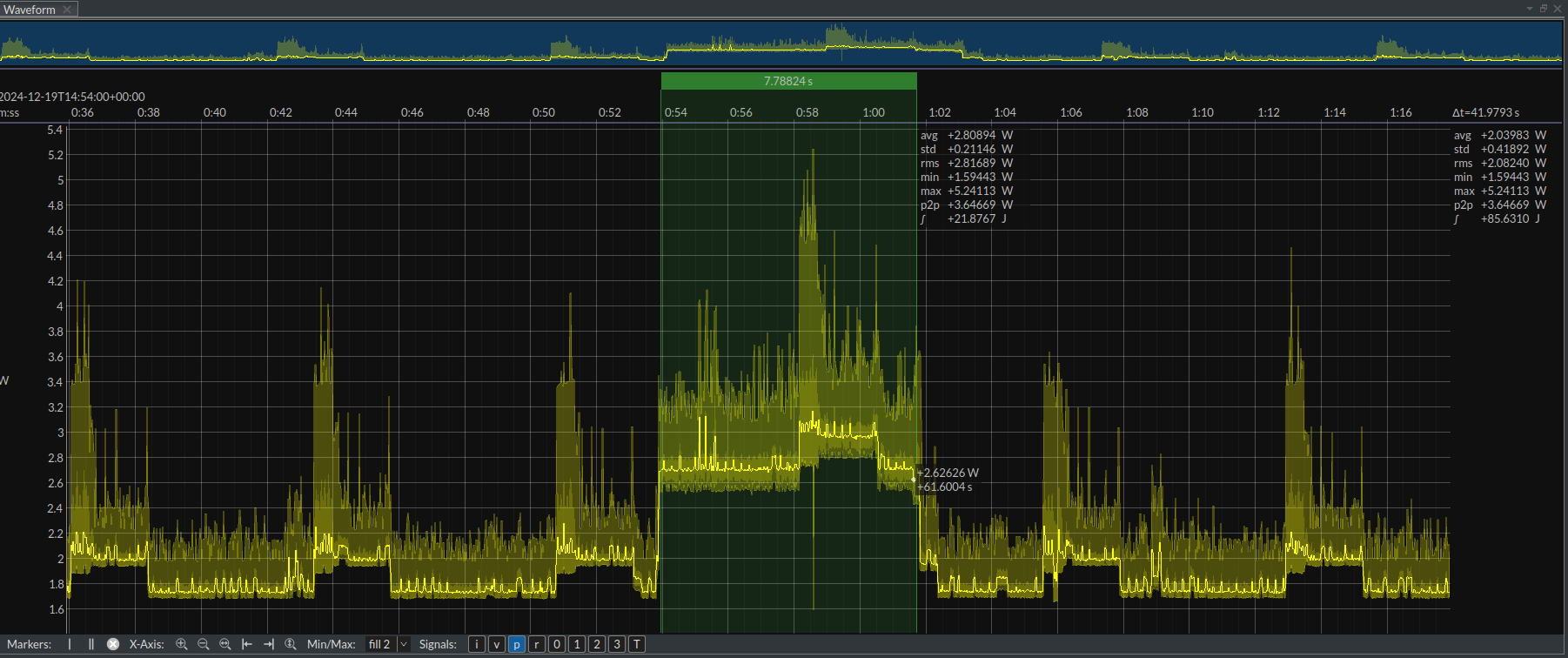}
\caption{Integer-only implementation}
\label{fig:power_int}
\end{subfigure}

\caption{The power consumption profile of a floating-point and integer-only implementation of 14,500,000 inferences of a Random Forest model with 50 classifiers and a maximum depth of 7.}
\label{fig:power_measurements}
\end{figure}

Since the granularity of the operation is large enough, we visually define the region of interest for the measurements. We started by establishing a baseline, by measuring the power consumption of the Raspberry Pi with no application running. The baseline measurements are shown in \autoref{fig:power_baseline}. The baseline power consumption is around 1.67 Watts, but there is a periodic background process that raises the average power consumption momentarily to just under 2 Watts. This gives us an average baseline power consumption of about 1.82 Watts.

Figures \ref{fig:power_float} and \ref{fig:power_int} show the power consumption profile for the floating-point implementation and the integer-only implementation, respectively. We observe a very slight decrease in power consumption for the integer-only implementation; however, this difference is not statistically significant. For both implementations, the average power consumption is 2.81 watts, with the runtime being 19.36 seconds for the floating-point implementation and 7.79 seconds for the integer implementation.

We can calculate the percentage of energy saved by dividing the energy spending for the integer implementation by the energy spending of the floating-point implementation over the same time, and subtracting this from 1. This can be done with the following formula:

$$E_{saved} = 1 - \frac{T_{int}\cdot P_{high}+(T_{float}-T_{int})\cdot P_{low}}{T_{float}\cdot P_{high}}$$

where $T_{int}$ is the runtime of the integer implementation in seconds, $P_{high}$ is the power consumption of the device when running an application in Watts, $T_{float}$ is the runtime of the floating-point implementation in seconds, and $P_{low}$ is the baseline power consumption of the device in Watts.

For this experiment, this results in:

$$E_{saved} = 1 - \frac{7.79\cdot 2.81+(19.36-7.79)\cdot 1.81}{19.36\cdot 2.81} \approx 0.213$$

We see a 21.3\% decrease in energy consumption for the integer implementation. However, in this scenario, the baseline power consumption is relatively high, due to the background processes running on the Raspberry Pi. This results in a smaller relative difference between the power consumption when running the inference compared to the baseline, limiting the potential energy savings. In real-world scenarios where low energy consumption is critical, it is likely that there are fewer background processes, and the baseline power consumption is more limited.

To conclude, for this experiment, we did not observe any significant difference in power consumption between the floating-point implementation and the integer-only implementation. However, since the integer-only implementation runs faster than the floating-point implementation, the energy consumption decreased by approximately 21.3\% for the same workload. However, we expect that in a more optimized environment, where high energy efficiency is critical, this value will be closer to 50\%, as background processes might be optimized to decrease the baseline power consumption. Furthermore, the ability to use tree-based models on lower-power devices without needing FPUs will lead to even higher energy savings, as the use of a much smaller core, without an FPU or its associated register file, can significantly reduce energy consumption.

\section{Conclusion}
\label{sec:Conclusion}
In this work, we propose InTreeger, an end-to-end framework for deploying integer-only architecture-agnostic decision trees.
We experimentally demonstrated InTreeger's capabilities to create decision trees ensembles for inference on hardware without FPUs, enabling the use of these devices for RFs and other tree-based machine learning models.

Additionally, we show that InTreeger can achieve notable performance improvements across three distinct architectures. These improvements are somewhat influenced by the number of classes within the dataset. 
While it is clear that the ability to exclude FPUs for decision tree inference will benefit the energy consumption of machine learning on the edge, we also conducted energy measurements that show a significant improvement in energy efficiency, even for chips with a relatively high baseline power consumption. All these benefits come for free, as we demonstrated that there is no loss of accuracy.

Furthermore, by releasing a framework that automizes the steps needed for training and deploying highly optimized tree based machine learning models, we enable anyone to easily apply these models anywhere, as the generated model is completely architecture agnostic.



\bibliographystyle{preamble/IEEEtran}
\bibliography{refs}

\begin{thebibliography}{10}
\providecommand{\url}[1]{#1}
\csname url@samestyle\endcsname
\providecommand{\newblock}{\relax}
\providecommand{\bibinfo}[2]{#2}
\providecommand{\BIBentrySTDinterwordspacing}{\spaceskip=0pt\relax}
\providecommand{\BIBentryALTinterwordstretchfactor}{4}
\providecommand{\BIBentryALTinterwordspacing}{\spaceskip=\fontdimen2\font plus
\BIBentryALTinterwordstretchfactor\fontdimen3\font minus
  \fontdimen4\font\relax}
\providecommand{\BIBforeignlanguage}[2]{{%
\expandafter\ifx\csname l@#1\endcsname\relax
\typeout{** WARNING: IEEEtran.bst: No hyphenation pattern has been}%
\typeout{** loaded for the language `#1'. Using the pattern for}%
\typeout{** the default language instead.}%
\else
\language=\csname l@#1\endcsname
\fi
#2}}
\providecommand{\BIBdecl}{\relax}
\BIBdecl

\bibitem{SHWARTZZIV202284}
R.~Shwartz-Ziv and A.~Armon, ``Tabular data: Deep learning is not all you
  need,'' \emph{Information Fusion}, vol.~81, pp. 84--90, 2022.

\bibitem{NEURIPS2018_14491b75}
L.~Prokhorenkova, G.~Gusev, A.~Vorobev, A.~V. Dorogush, and A.~Gulin,
  ``Catboost: unbiased boosting with categorical features,'' in \emph{Advances
  in Neural Information Processing Systems}, S.~Bengio, H.~Wallach,
  H.~Larochelle, K.~Grauman, N.~Cesa-Bianchi, and R.~Garnett, Eds., vol.~31,
  2018.

\bibitem{Pedretti2021}
G.~Pedretti, C.~E. Graves, S.~Serebryakov, R.~Mao, X.~Sheng, M.~Foltin, C.~Li,
  and J.~P. Strachan, ``Tree-based machine learning performed in-memory with
  memristive analog cam,'' \emph{Nature Communications}, vol.~12, no.~1, p.
  5806, Oct 2021.

\bibitem{9923840}
A.~Prasad, S.~Rajendra, K.~Rajan, R.~Govindarajan, and U.~Bondhugula,
  ``Treebeard: An optimizing compiler for decision tree based ml inference,''
  in \emph{55th IEEE/ACM International Symposium on Microarchitecture (MICRO)},
  2022, pp. 494--511.

\bibitem{10.5555/3294996.3295074}
G.~Ke, Q.~Meng, T.~Finley, T.~Wang, W.~Chen, W.~Ma, Q.~Ye, and T.-Y. Liu,
  ``Lightgbm: a highly efficient gradient boosting decision tree,'' in
  \emph{Proceedings of the 31st International Conference on Neural Information
  Processing Systems}, 2017, p. 3149–3157.

\bibitem{10.1145/3508019}
K.-H. Chen, C.~Su, C.~Hakert, S.~Buschj\"{a}ger, C.-L. Lee, J.-K. Lee,
  K.~Morik, and J.-J. Chen, ``Efficient realization of decision trees for
  real-time inference,'' \emph{ACM Trans. Embed. Comput. Syst.}, vol.~21,
  no.~6, 2022.

\bibitem{10153433}
F.~Daghero, A.~Burrello, E.~Macii, P.~Montuschi, M.~Poncino, and
  D.~Jahier~Pagliari, ``Dynamic decision tree ensembles for energy-efficient
  inference on iot edge nodes,'' \emph{IEEE Internet of Things Journal},
  vol.~11, no.~1, pp. 742--757, 2024.

\bibitem{10.1145/3704727}
\BIBentryALTinterwordspacing
C.~Su, C.-H. Ku, J.~K. Lee, and K.-H. Chen, ``Treehouse: An mlir-based
  compilation flow for real-time tree-based inference,'' \emph{ACM Trans.
  Embed. Comput. Syst.}, Nov. 2024. [Online]. Available:
  \url{https://doi.org/10.1145/3704727}
\BIBentrySTDinterwordspacing

\bibitem{ULLAH2023116614}
Z.~Ullah, N.~Yoon, B.~K. Tarus, S.~Park, and M.~Son, ``Comparison of tree-based
  model with deep learning model in predicting effluent ph and concentration by
  capacitive deionization,'' \emph{Desalination}, vol. 558, p. 116614, 2023.

\bibitem{inemo}
``{iNEMO inertial module: Always-on 3D accelerometer and 3D gyro-scope},'' Data
  Sheet, STMicroelectronics, 2019,
  https://www.st.com/resource/en/datasheet/lsm6dsox.pdf.

\bibitem{BeniniPaper}
E.~Tabanelli, G.~Tagliavini, and L.~Benini, ``Optimizing random forest-based
  inference on risc-v mcus at the extreme edge,'' \emph{IEEE Transactions on
  Computer-Aided Design of Integrated Circuits and Systems}, vol.~41, no.~11,
  pp. 4516--4526, 2022.

\bibitem{10.1023/A:1010933404324}
L.~Breiman, ``Random forests,'' \emph{Mach. Learn.}, vol.~45, no.~1, p. 5–32,
  oct 2001.

\bibitem{Geurts2006}
P.~Geurts, D.~Ernst, and L.~Wehenkel, ``Extremely randomized trees,''
  \emph{Machine Learning}, vol.~63, no.~1, pp. 3--42, Apr 2006.

\bibitem{10.1214/aos/1013203451}
J.~H. Friedman, ``{Greedy function approximation: A gradient boosting
  machine.}'' \emph{The Annals of Statistics}, vol.~29, no.~5, pp. 1189 --
  1232, 2001.

\bibitem{scikit-learn}
F.~Pedregosa, G.~Varoquaux, A.~Gramfort, V.~Michel, B.~Thirion, O.~Grisel,
  M.~Blondel, P.~Prettenhofer, R.~Weiss, V.~Dubourg, J.~Vanderplas, A.~Passos,
  D.~Cournapeau, M.~Brucher, M.~Perrot, and E.~Duchesnay, ``Scikit-learn:
  Machine learning in {P}ython,'' \emph{Journal of Machine Learning Research},
  vol.~12, pp. 2825--2830, 2011.

\bibitem{randomForestR}
\BIBentryALTinterwordspacing
A.~Liaw and M.~Wiener, ``Classification and regression by randomforest,''
  \emph{R News}, vol.~2, no.~3, pp. 18--22, 2002. [Online]. Available:
  \url{https://CRAN.R-project.org/doc/Rnews/}
\BIBentrySTDinterwordspacing

\bibitem{10.1145/3652032.3659366}
J.~Malcher, D.~Biebert, K.-H. Chen, S.~Buschj\"{a}ger, C.~Hakert, and J.-J.
  Chen, ``Language-based deployment optimization for random forests (invited
  paper),'' in \emph{Proceedings of the 25th ACM SIGPLAN/SIGBED International
  Conference on Languages, Compilers, and Tools for Embedded Systems}, 2024, p.
  58–61.

\bibitem{JSSv077i01}
M.~N. Wright and A.~Ziegler, ``ranger: A fast implementation of random forests
  for high dimensional data in c++ and r,'' \emph{Journal of Statistical
  Software}, vol.~77, no.~1, p. 1–17, 2017.

\bibitem{6513227}
N.~Asadi, J.~Lin, and A.~P. de~Vries, ``Runtime optimizations for tree-based
  machine learning models,'' \emph{IEEE Transactions on Knowledge and Data
  Engineering}, vol.~26, no.~9, pp. 2281--2292, 2014.

\bibitem{DBLP:conf/icdm/BuschjagerCCM18}
S.~Buschj{\"{a}}ger, K.~Chen, J.~Chen, and K.~Morik, ``Realization of random
  forest for real-time evaluation through tree framing,'' in \emph{{IEEE}
  International Conference on Data Mining, {ICDM} 2018, Singapore, November
  17-20, 2018}, 2018, pp. 19--28.

\bibitem{6239820}
B.~Van~Essen, C.~Macaraeg, M.~Gokhale, and R.~Prenger, ``Accelerating a random
  forest classifier: Multi-core, gp-gpu, or fpga?'' in \emph{IEEE 20th
  International Symposium on Field-Programmable Custom Computing Machines},
  2012, pp. 232--239.

\bibitem{7962153}
S.~Buschjäger and K.~Morik, ``Decision tree and random forest implementations
  for fast filtering of sensor data,'' \emph{IEEE Transactions on Circuits and
  Systems I: Regular Papers}, vol.~65, no.~1, pp. 209--222, 2018.

\bibitem{10247695}
C.-L. Tsai, C.-F. Wu, Y.-H. Chang, H.-W. Hu, Y.-C. Lee, H.-P. Li, and T.-W.
  Kuo, ``A digital 3d tcam accelerator for the inference phase of random
  forest,'' in \emph{60th ACM/IEEE Design Automation Conference (DAC)}, 2023,
  pp. 1--6.

\bibitem{10.1007/978-3-031-26419-1_32}
C.~Hakert, K.-H. Chen, and J.-J. Chen, ``Immediate split trees: Immediate
  encoding of floating point split values in random forests,'' in
  \emph{Machine Learning and Knowledge Discovery in Databases}, M.-R. Amini,
  S.~Canu, A.~Fischer, T.~Guns, P.~Kralj~Novak, and G.~Tsoumakas, Eds., 2023,
  pp. 531--546.

\bibitem{DBLP:journals/tecs/ChenSHBLLMC22}
K.~Chen, C.~Su, C.~Hakert, S.~Buschj{\"{a}}ger, C.~Lee, J.~Lee, K.~Morik, and
  J.~Chen, ``Efficient realization of decision trees for real-time inference,''
  \emph{{ACM} Trans. Embed. Comput. Syst.}, vol.~21, no.~6, pp. 68:1--68:26,
  2022.

\bibitem{10546851}
C.~Hakert, K.-H. Chen, and J.-J. Chen, ``Flint: Exploiting floating point
  enabled integer arithmetic for efficient random forest inference,'' in
  \emph{Design, Automation \& Test in Europe Conference \& Exhibition}, 2024,
  pp. 1--2.

\bibitem{DBLP:conf/mlsys/2018}
H.~Cho and M.~Li, ``Treelite: toolbox for decision tree deployment,'' in
  \emph{Proceedings of Machine Learning and Systems (MLSys)}, 2018.

\bibitem{DBLP:conf/cvpr/JacobKCZTHAK18}
B.~Jacob, S.~Kligys, B.~Chen, M.~Zhu, M.~Tang, A.~G. Howard, H.~Adam, and
  D.~Kalenichenko, ``Quantization and training of neural networks for efficient
  integer-arithmetic-only inference,'' in \emph{2018 {IEEE} Conference on
  Computer Vision and Pattern Recognition, {CVPR} 2018, Salt Lake City, UT,
  USA, June 18-22, 2018}, 2018, pp. 2704--2713.

\bibitem{XGBoost}
T.~Chen and C.~Guestrin, ``Xgboost: A scalable tree boosting system,'' in
  \emph{Proceedings of the 22nd ACM SIGKDD International Conference on
  Knowledge Discovery and Data Mining}, 2016, p. 785–794.

\bibitem{10.1007/978-3-030-46150-8_35}
L.~Devos, W.~Meert, and J.~Davis, ``Fast gradient boosting decision trees with
  bit-level data structures,'' in \emph{Machine Learning and Knowledge
  Discovery in Databases}, U.~Brefeld, E.~Fromont, A.~Hotho, A.~Knobbe,
  M.~Maathuis, and C.~Robardet, Eds., 2020, pp. 590--606.

\bibitem{8766229}
``Ieee standard for floating-point arithmetic,'' \emph{IEEE Std 754-2019
  (Revision of IEEE 754-2008)}, pp. 1--84, 2019.

\bibitem{oshiro2012many}
T.~M. Oshiro, P.~S. Perez, and J.~A. Baranauskas, ``How many trees in a random
  forest?'' in \emph{Machine Learning and Data Mining in Pattern Recognition:
  8th International Conference, MLDM 2012, Berlin, Germany, July 13-20, 2012.
  Proceedings 8}.\hskip 1em plus 0.5em minus 0.4em\relax Springer, 2012, pp.
  154--168.

\bibitem{joulescope_js220}
\BIBentryALTinterwordspacing
{Joulescope}, ``Js220 joulescope precision energy analyzer,'' 2025, accessed:
  2025-01-21. [Online]. Available:
  \url{https://www.joulescope.com/products/js220-joulescope-precision-energy-analyzer}
\BIBentrySTDinterwordspacing

\bibitem{shuttle_dataset}
``{Statlog (Shuttle)},'' UCI Machine Learning Repository, {DOI}:
  https://doi.org/10.24432/C5WS31.

\bibitem{esa_dataset}
\BIBentryALTinterwordspacing
G.~De~Canio, K.~Kotowski, and C.~Haskamp, ``\BIBforeignlanguage{en}{Esa anomaly
  dataset},'' Jun. 2024. [Online]. Available:
  \url{https://zenodo.org/doi/10.5281/zenodo.12528696}
\BIBentrySTDinterwordspacing

\bibitem{joulescope_man}
{Joulescope}, \emph{\BIBforeignlanguage{English}{Joulescope™ JS220 User’s
  Guide Precision DC Energy Analyzer}}, {Joulescope}.

\end{thebibliography}

\end{document}